\documentclass[letterpaper]{article} 
\usepackage{aaai2027}
\usepackage[hyphens]{url}  
\usepackage{graphicx} 
\usepackage{natbib}  
\usepackage{caption} 
\usepackage{algorithm}
\usepackage{algorithmic}
\usepackage{booktabs}
\usepackage{amsmath}
\usepackage{amsthm}
\usepackage{subcaption}
\usepackage{amssymb}
\usepackage{multirow}
\usepackage[table]{xcolor}
\usepackage{makecell}

\definecolor{LightBlue}{RGB}{232,244,255}
\definecolor{SoftYellow}{RGB}{250,245,210}

\usepackage{newfloat}
\usepackage{listings}
\DeclareCaptionStyle{ruled}{labelfont=normalfont,labelsep=colon,strut=off} 
\floatstyle{ruled}
\newfloat{listing}{tb}{lst}{}
\floatname{listing}{Listing}

\usepackage{booktabs}

\title{Trajectory-Guided Forget-Recover Network for Continual LLM Unlearning}
 \author{
    Zezheng Wu\textsuperscript{\rm 1},
    Xinghe Cheng\textsuperscript{\rm 2},
    Qinggang Zhang\textsuperscript{\rm 3},
    Haoran Luo\textsuperscript{\rm 4},\\
    Jiapu Wang\textsuperscript{\rm 5},
    Qing Yang\textsuperscript{\rm 1},
    Jingwei Zhang\textsuperscript{\rm 1}
 }
\affiliations{
    \textsuperscript{\rm 1}Guilin University of Electronic Technology\quad
    \textsuperscript{\rm 2}Jinan University\quad
    \textsuperscript{\rm 3}Jilin University\\
    \textsuperscript{\rm 4}Nanyang Technological University\quad
    \textsuperscript{\rm 5}Nanjing University of Science and Technology\\

    wzz@mails.guet.edu.cn, gtzjw@hotmail.com
}

\begin{document}

\maketitle

\begin{abstract}
Machine unlearning aims to eliminate the influence of sensitive data on a model. 
In the real world, unlearning requests arrive continually, which gives rise to two challenges.
First, an unlearning intervention may redistribute target-related computation across remaining pathways, allowing previously forgotten knowledge to re-emerge. Second, repeated unlearning interventions may progressively reduce the model capacity needed to preserve retained utility.
To address these challenges, we propose the Trajectory-guided Forget-Recover Network (TFR-Net). TFR-Net tracks channel-level risk across requests. It separates persistent target-related channels from transient hotspots and suppresses only the persistent ones. TFR-Net also recovers model capacity by reactivating dormant channels. These channels make strong contributions to retained utility and show low current and historical forget risk. The recovery is accepted only when retained-utility degradation remains within a predefined tolerance.
Experiments on four datasets show that TFR-Net consistently achieves a more favorable trade-off between unlearning effectiveness and retained utility than representative baselines.
\end{abstract}


\section{Introduction}
Machine unlearning aims to remove the influence of private, copyrighted, or obsolete information from Large Language Models (LLMs), providing a key safeguard for responsible deployment~\cite{YaoXL2024,YaoCD2024,cheng2026graphrag,wang2024large}. For deployed LLMs, unlearning is inherently continual because new requests may arise throughout the model lifecycle and must be fulfilled without undoing prior forgetting or impairing retained capabilities. Each new request acts on a model already changed by earlier unlearning. Remaining pathways may then take over the suppressed target-related computation, making previously forgotten knowledge accessible again. Meanwhile, repeated suppression leaves fewer usable pathways for preserving retained capabilities.

\begin{figure}[t]
\centering

\begin{subfigure}{\columnwidth}
    \centering
    \includegraphics[width=\linewidth]{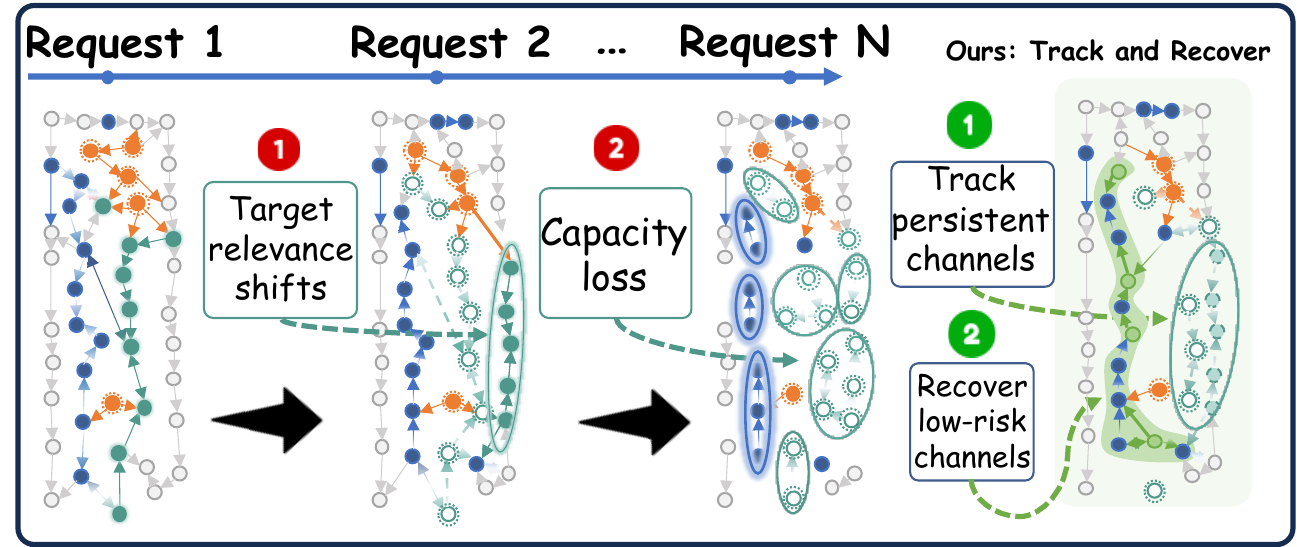}
    \caption{Continual unlearning challenges.}
    \label{fig:motivation_challenge}
\end{subfigure}

\vspace{0.6em}

\begin{subfigure}{\columnwidth}
    \centering
    \includegraphics[width=\linewidth]{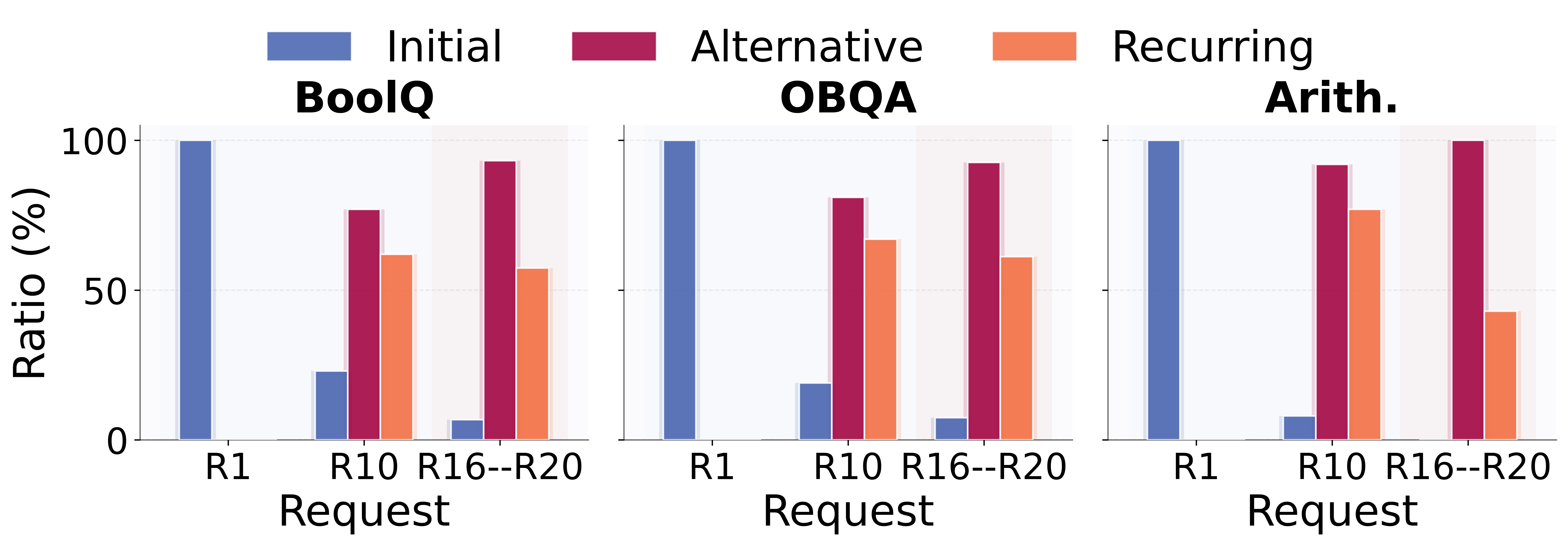}
    \caption{Structural evidence.}
    \label{fig:channel_evolution}
\end{subfigure}

\caption{
Motivation and evidence for TFR-Net.
(a) Repeated unlearning requests trigger self-repair and cumulative pathway damage.
(b) Forget-relevant pathways are progressively redistributed across rounds.
}
\label{fig:motivation_evolution}
\end{figure}

Existing continual unlearning methods mainly stabilize updates across successive forget requests by limiting inter-request interference~\cite{GaoWD2025,ZadZL2026,zhuo2025effective,yan2026cage}. However, suppressing the channels identified for the current request does not guarantee lasting forgetting. As the model changes under later requests, other active channels may begin to carry the same target-related computation. Although target-related units can be identified within a given model state~\cite{ZhaZZ2025,yan2025prompting}, their evolution across requests is rarely tracked. This limitation motivates tracking how target relevance evolves across requests.

Figure~\ref{fig:channel_evolution} shows that the Top-$K$ channels associated with target relevance change markedly across forget requests. By Req.~16--Req.~20, Initial channels account for only a small share across all datasets, whereas Alternative channels dominate. Many Alternative channels also reappear in later requests, as captured by the Recurring group. Thus, target relevance neither remains confined to a fixed initial set nor moves only through isolated one-request channels, but shifts among channels with different levels of persistence. The key challenge is therefore to distinguish persistent target-related channels from transient hotspots across requests.

A second challenge concerns the cumulative cost of repeated suppression. Each request leaves more channels suppressed, so later requests must operate with less usable model capacity. Because some of these channels also contribute to retained utility, accumulated dormancy progressively narrows the active channel space needed to maintain retained utility. Recovering dormant channels can restore capacity, but recovery cannot simply reverse previous suppression. The key challenge is therefore to recover useful capacity without compromising current or previously achieved forgetting.

To address these challenges, we propose the Trajectory-guided Forget-Recover Network (TFR-Net), which maintains persistent channel masks across forget requests. For each request, TFR-Net contrasts the channel score measured on the forget request and retain data, and summarizes the temporal evolution of channel risk in a cross-request trajectory. These trajectories enable selective suppression of persistent target-related channels while discounting transient hotspots. TFR-Net further employs retain-guarded capacity recovery, which ranks dormant channels by their contributions to retained utility under current and historical forget-risk constraints and partially reactivates eligible channels. The recovered model state is accepted only when the degradation in retained utility remains within tolerance; otherwise, the complete pre-request state is restored. Together, these mechanisms coordinate persistent target suppression with capacity preservation throughout the request stream.

Our contributions are summarized as follows:
\begin{itemize}
    \item We formulate continual unlearning for LLMs as the joint control of evolving target relevance and cumulative capacity loss, and characterize their structural patterns across requests.

    \item We propose TFR-Net, which uses persistent channel masks and cross-request risk trajectories to identify persistent target-related channels and distinguish them from transient hotspots.

    \item We introduce retain-guarded capacity recovery, which selectively reactivates dormant channels based on retained utility and current and historical forget risk.

    \item Extensive experiments on four datasets show that TFR-Net consistently improves the forgetting--utility balance, achieving the highest Trade-off on all three continual unlearning streams.
\end{itemize}

\section{Related Work}
\label{relaed}
\textbf{Machine Unlearning}
Machine Unlearning (MU) originated from regulatory mandates such as the "right to be forgotten"~\cite{CaoY2015, BourtouleCC2021}, aiming to eliminate the influence of specific training points from model parameters~\cite{NguyenHR2025, ZhangHB2024, DeHP2023, LiWY2024, 0012WYY0Y25, 0012YP24, ZhangWYPTP24}. Early work focused on discriminative tasks~\cite{GinartGV2019}, while recent studies increasingly address generative unlearning in LLMs to mitigate privacy leakage risks~\cite{JangYY2023, CarliniTW2021}. Primary unlearning techniques include Gradient Ascent (GA), which is often hindered by hyperparameter sensitivity and catastrophic forgetting~\cite{JangYY2023, FanLZ2023}; preference-based approaches like SimNPO~\cite{FaLL2026}; and strategies relying on random labeling or mismatch objectives~\cite{LiuYJ2025}. Recent advancements emphasize localization-based paradigms~\cite{LiuYJ2025, LiCZ2024}, which identify and modify specific computational units responsible for target knowledge~\cite{WuLX2023, FanLZ2023}. Building upon this, frameworks such as LLM-Eraser utilize selective pruning and contrastive distillation to effectively disentangle undesired knowledge while preserving general model performance~\cite{ZhaZZ2025}.

\noindent\paragraph{\textbf{Continual Machine Unlearning}}
Continual machine unlearning extends one-shot unlearning to a sequence of forget requests. Existing methods mainly improve this process by stabilizing updates across requests. O3 isolates request-specific changes through orthogonal low-rank adapters~\cite{GaoWD2025}, while ALKN adapts localization and unlearning strength to preserve utility under continual requests~\cite{AbuWC2025}. ASU further improves robustness through attention-smoothed self-distillation~\cite{ZadZL2026}. These methods primarily improve update stability across successive forget requests, while the cross-request evolution of target-related computation remains less explicitly modeled. Consequently, the persistence of target-related channels and the cumulative depletion of usable model capacity under repeated suppression remain less directly addressed.

\begin{figure*}[ht]
  \centering
  \includegraphics[width=1.0\textwidth]{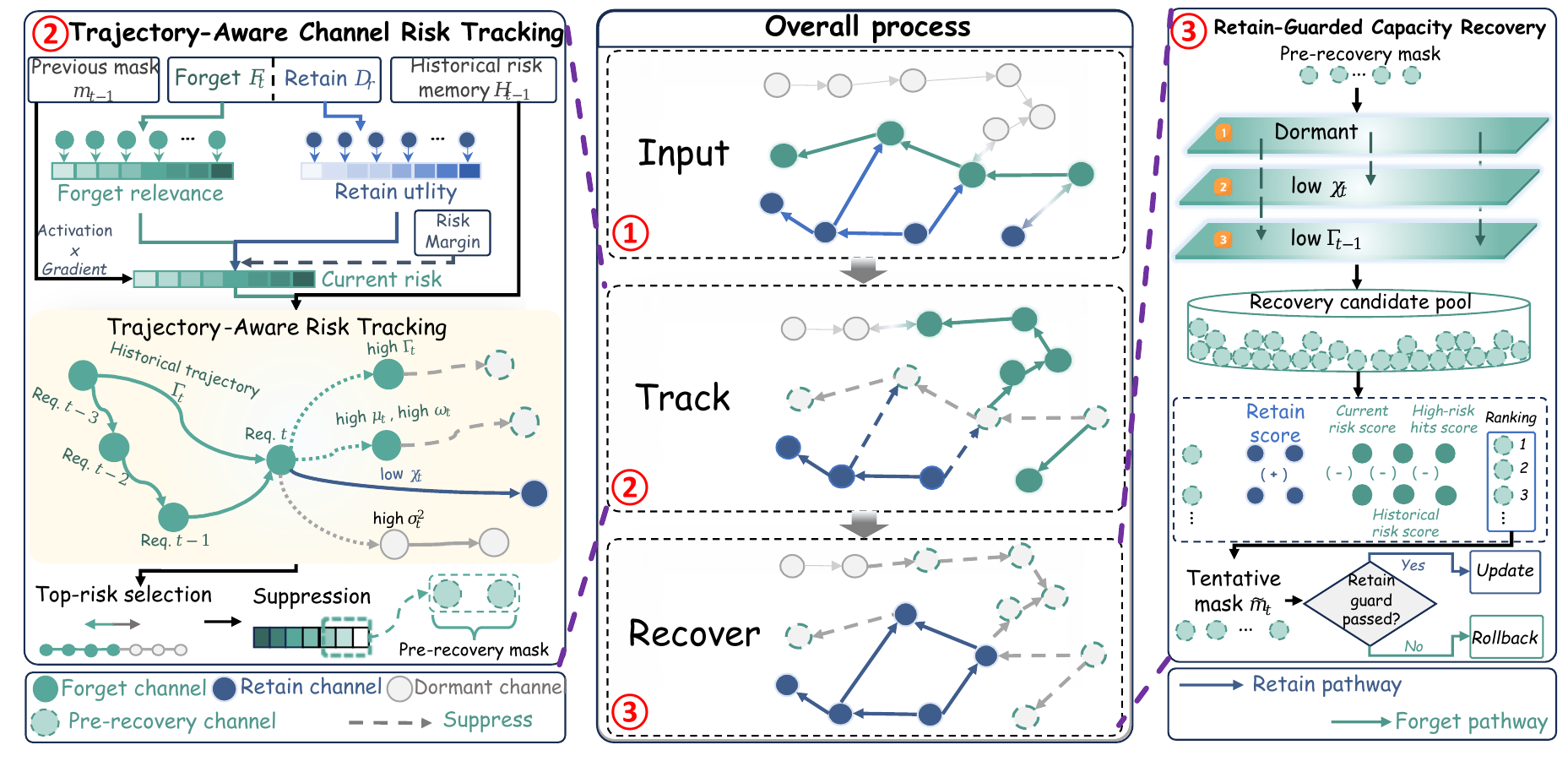}
    \caption{
    Overview of TFR-Net.
    TFR-Net performs continual unlearning through persistent pathway masking, self-repair pathway tracking, and risk-safe pathway reactivation.
    It first maintains evolving masks across continual forget requests, then identifies self-repaired forget pathways using forget-retain signal contrast and trajectory risk memory.
    High-priority risks are selectively suppressed to weaken stable forget pathways, while low-risk dormant candidates are safely reactivated for capacity recovery.
    An atomic retain guard commits or restores the complete structural state.
    }
  \label{fig:png_main}
\end{figure*}

\section{Problem Definition}

\noindent \textbf{Request stream.}
We study continual unlearning for LLMs under a stream of forget requests
\begin{equation}
\mathcal{S}_f=\{\mathcal{F}_t\}_{t=1}^{T},
\label{eq:forget_stream}
\end{equation}
where $\mathcal{F}_t$ contains the target samples specified for unlearning in request $t$, and $\mathcal{D}_r$ is the retain set. Each request acts on a model modified by previous requests, requiring sustained unlearning effectiveness and retained utility.

\noindent \paragraph{\textbf{Mask trajectory.}}
Let $\mathcal{C}$ denote the candidate channels in selected transformer projections. TFR-Net represents evolving pathway states by the accepted mask $\mathbf{m}_t=(m_t(c))_{c\in\mathcal{C}}$, where $m_t(c)\in[0,1]$, $\mathbf{m}_0=\mathbf{1}$, and the masked model is $f_{\theta_0,\mathbf{m}_t}$. Smaller values reduce channel contributions, and channel $c$ is dormant when $m_t(c)\leq\tau_d$. The online update is
\begin{equation}
\mathbf{m}_t=\pi(\mathcal{F}_t,\mathcal{D}_r,\mathbf{m}_{t-1},\mathcal{H}_{t-1}),
\label{eq:online_mask_update}
\end{equation}
where $\pi$ is the mask-update rule and $\mathcal{H}_{t-1}$ stores channel-level risk statistics accumulated before request $t$.

\noindent \paragraph{\textbf{Objective.}}
The mask trajectory balances current and historical unlearning effectiveness, retained utility, and usable model capacity. A conceptual cumulative objective is
\begin{equation}
\begin{aligned}
\min_{\pi}\sum_{t=1}^{T}\Big[
&\mathcal{R}_f(f_{\theta_0,\mathbf{m}_t};\mathcal{F}_t)
+\lambda_h\mathcal{R}_h(f_{\theta_0,\mathbf{m}_t};\mathcal{H}_{t-1})\\
&+\lambda_r[\Delta_r(t)]_+
+\lambda_s\mathrm{Sp}(\mathbf{m}_t)
\Big],
\end{aligned}
\label{eq:continual_unlearning_objective}
\end{equation}
where $\mathcal{R}_f$ and $\mathcal{R}_h$ denote current-request residual risk and historical forget risk, $\Delta_r(t)$ denotes retained-utility degradation, and $\mathrm{Sp}(\mathbf{m}_t)$ denotes the dormant-channel ratio.

\section{Methodology}
Figure~\ref{fig:png_main} illustrates TFR-Net. \textbf{Persistent Channel Masking} maintains the evolving structural state across requests. \textbf{Trajectory-Aware Channel Risk Tracking} identifies persistent target-related channels for selective suppression. \textbf{Retain-Guarded Capacity Recovery} safely reactivates low-risk dormant channels. Together, these components improve the long-horizon forgetting--utility trade-off.

\subsection{Persistent Channel Masking}
TFR-Net represents pathway evolution across requests with persistent channel masks. It keeps the pretrained backbone parameters fixed and updates masks over selected projection channels.

Let each candidate channel be indexed by $c=(l,g,j)$, where $l$ denotes the layer, $g$ denotes the projection type, and $j$ denotes the output channel. We use $m_t(c)$ as the compact notation and $m_t^{l,g,j}$ as its layer-wise expanded form. Consistent with Problem Definition, the structural mask at request $t$ is defined as follows:
\begin{equation}
\mathbf{m}_t
=
\big(m_t(c)\big)_{c\in\mathcal{C}},
\label{eq:structural_mask}
\end{equation}
where $m_t(c)\in[0,1]$ controls the contribution of channel $c$. The initial mask $\mathbf{m}_0$ activates all candidate channels before the first forget request.

Given hidden states $\mathbf{X}^{l}$, TFR-Net applies the channel mask to the output of projection $g$ at layer $l$ as follows:
\begin{equation}
\widetilde{H}_{t,b,n,j}^{l,g}
=
\left(
\sum_{d=1}^{d_l}
X_{t,b,n,d}^{l} W_{j,d}^{l,g}
+
b_{j}^{l,g}
\right)
m_{t}^{l,g,j},
\label{eq:masked_projection}
\end{equation}
where $\mathbf{W}^{l,g}$ and $\mathbf{b}^{l,g}$ are frozen projection parameters, $b\in\{1,\ldots,B\}$ indexes the batch dimension, $n\in\{1,\ldots,N\}$ indexes the token position, $j\in\{1,\ldots,d_g\}$ indexes the output channel, and $m_t^{l,g,j}\in[0,1]$ is the mask value applied to the $j$-th projection channel. A smaller mask value suppresses the contribution of the corresponding projection channel, while a larger value preserves its role in forward computation.

The effective masked model after the $t$-th forget request is written as follows:
\begin{equation}
f_t(x)
=
f_{\theta_0,\mathbf{m}_t}(x),
\label{eq:effective_masked_model}
\end{equation}
where $\theta_0$ denotes the frozen backbone parameters. Under this formulation, continual unlearning is implemented by evolving the mask trajectory rather than directly modifying the backbone.

To measure how much usable model capacity remains after structural suppression, TFR-Net counts the channels whose mask values stay above the dormant threshold as follows:
\begin{equation}
\begin{aligned}
\mathrm{Cap}(\mathbf{m}_t)
=
\frac{
\sum_{g}\sum_{l}\sum_{j=1}^{d_g}
\mathbb{I}\big[m_t^{l,g,j}>\tau_d\big]
}{
\sum_{g}|\Omega_g|d_g
},
\end{aligned}
\label{eq:structural_capacity}
\end{equation}
where $\tau_d$ is the dormant threshold, $\Omega_g$ denotes the selected layers for projection type $g$, and the summations are taken over selected projection types, selected layers, and output channels. $\mathrm{Cap}(\mathbf{m}_t)$ measures the fraction of channels still available for effective computation. A lower value indicates stronger accumulated channel dormancy and less usable capacity for subsequent forget requests. The dormant-channel ratio in Problem Definition is
$\mathrm{Sp}(\mathbf{m}_t)=1-\mathrm{Cap}(\mathbf{m}_t)$.

The suppression-recovery transition is defined as follows:
\begin{equation}
\begin{aligned}
\widetilde{m}_t^{l,g,j}
=
\mathrm{clip}_{[0,1]}
\big(
m_{t-1}^{l,g,j}\alpha_t^{l,g,j}
+
\eta_t^{l,g,j}
\big),
\end{aligned}
\label{eq:forget_recovery_transition}
\end{equation}
where $\alpha_t^{l,g,j}$ is the suppression factor determined by the trajectory-aware channel priority, and $\eta_t^{l,g,j}$ is the reactivation increment assigned to the selected recovery channels. 

\subsection{Trajectory-Aware Channel Risk Tracking}
This subsection models pathway evolution through channel-level risk trajectories and prioritizes active channels that remain persistently target-related.

For each candidate channel $c$, TFR-Net computes activation-gradient
scores on the forget request and retain data under the incoming mask
$\mathbf{m}_{t-1}$. Let $a_{t,c}^{i}(x)$ denote the pre-mask projection
activation of channel $c$ at token position $i$, and define the masked activation as:
\begin{equation}
\tilde{h}_{t,c}^{i}(x)
=
m_{t-1}(c)a_{t,c}^{i}(x),
\label{eq:masked_channel_activation}
\end{equation}
denote the corresponding masked activation. The channel score under data
split $q$ is defined as:
\begin{equation}
\begin{aligned}
\psi_t^{q}(c)
=
\frac{1}{|\mathcal{B}_q|}
\sum_{x\in\mathcal{B}_q}
\left|
\frac{1}{|\mathcal{T}_x|}
\sum_{i\in\mathcal{T}_x}
a_{t,c}^{i}(x)
\frac{\partial \mathcal{L}_q(x)}
{\partial \tilde{h}_{t,c}^{i}(x)}
\right|,
\end{aligned}
\label{eq:channel_score}
\end{equation}
where $q\in\{f,r\}$ denotes the forget or retain split,
$\mathcal{B}_q$ is the corresponding mini-batch, $\mathcal{T}_x$ is the
token set used for score aggregation, and $\mathcal{L}_q$ is the
corresponding loss. This score combines the pre-mask activation
$a_{t,c}^{i}(x)$ with the gradient with respect to the masked activation
$\tilde{h}_{t,c}^{i}(x)$, providing a token-aggregated first-order proxy
for the sensitivity of the channel mask. Consequently, the score remains
informative for dormant channels even when their current masked
contributions are small. $\mathcal{L}_f$ and $\mathcal{L}_r$ denote the corresponding task losses on the forget and retain batches, respectively.

The current-request channel risk is computed from the contrast between the
normalized forget and retain scores:
\begin{equation}
\chi_t(c)
=
\big[
\zeta_f \bar{\psi}_t^{f}(c)
-
\zeta_r \bar{\psi}_t^{r}(c)
-
\gamma
\big]_+,
\label{eq:current_pathway_risk}
\end{equation}
where $\bar{\psi}_t^{f}(c)$ and $\bar{\psi}_t^{r}(c)$ denote median and MAD standardized forget and retain scores within each module, $\zeta_f$ and $\zeta_r$ control their relative weights, $\gamma$ is a risk margin, and $[\cdot]_+$ denotes positive clipping. A large $\chi_t(c)$ indicates stronger relevance to the current forget request relative to its contribution to retained utility.

Single-request risk can vary as target relevance is redistributed across channels over successive requests. TFR-Net therefore maintains a temporal memory for each channel. The long-term risk level is updated as follows:
\begin{equation}
\mu_t(c)
=
\beta_\mu \mu_{t-1}(c)
+
(1-\beta_\mu)\chi_t(c),
\label{eq:risk_ema}
\end{equation}
where $\mu_t(c)$ is the smoothed risk memory and $\beta_\mu$ controls the memory decay. 

To capture temporal variation in channel risk, TFR-Net tracks the first-order risk change as follows:
\begin{equation}
\nu_t(c)
=
\mu_t(c)-\mu_{t-1}(c),
\label{eq:risk_velocity}
\end{equation}
where $\nu_t(c)$ measures the risk change between two consecutive requests. A large magnitude indicates that the channel risk is rapidly changing and may be less stable.

TFR-Net further tracks the second-order change as follows:
\begin{equation}
\xi_t(c)
=
\nu_t(c)-\nu_{t-1}(c),
\label{eq:risk_acceleration}
\end{equation}
where $\xi_t(c)$ captures abrupt second-order variation in channel risk. 
This term reduces the priority of channels with abrupt short-term risk fluctuations.

The temporal uncertainty of the channel risk is estimated as follows:
\begin{equation}
\sigma_t^2(c)
=
\beta_\sigma\sigma_{t-1}^2(c)
+
(1-\beta_\sigma)\big(\chi_t(c)-\mu_t(c)\big)^2,
\label{eq:risk_variance}
\end{equation}
where $\sigma_t^2(c)$ measures the variance of current risk around the smoothed risk memory. High uncertainty reduces confidence that a channel is persistently target-related.

To emphasize channels that repeatedly appear in the high-risk region, TFR-Net maintains a high-risk hit memory as follows:
\begin{equation}
\omega_t(c)
=
\beta_\omega\omega_{t-1}(c)
+
(1-\beta_\omega)\mathbb{I}\big[\chi_t(c)>\kappa_t\big],
\label{eq:risk_hit_memory}
\end{equation}
where $\omega_t(c)$ records repeated high-risk occurrences and $\kappa_t$ is the current-request risk threshold.

The trajectory-aware pathway priority is then defined as follows:
\begin{equation}
\begin{aligned}
\Gamma_t(c)
=
\frac{
\mu_t(c)\big(1+\lambda_\omega\omega_t(c)\big)
}{
1+
|\nu_t(c)|
+
|\xi_t(c)|
+
\sqrt{\sigma_t^2(c)}
},
\end{aligned}
\label{eq:trajectory_priority}
\end{equation}
where $\Gamma_t(c)$ is the final priority for suppressing channel $c$. The numerator emphasizes persistent and repeatedly observed forget risk, while the denominator penalizes unstable risk change and temporal uncertainty.

Finally, TFR-Net converts trajectory priority into a suppression factor for mask update. Let $\mathcal{P}_t$ denote the selected high-priority active channels under the suppression budget. The suppression factor is defined as follows:
\begin{equation}
\alpha_t(c)
=
1-\delta_t\mathbb{I}\big[c\in\mathcal{P}_t\big],
\label{eq:suppression_factor}
\end{equation}
where $\delta_t$ controls the suppression strength. When $c=(l,g,j)$, $\alpha_t(c)$ corresponds to $\alpha_t^{l,g,j}$ in the forget-recover transition. During suppression, channels in $\mathcal{P}_t$ are multiplicatively reduced by $1-\delta_t$, while the other active channels retain their mask values.

\begin{table*}[t]
\tiny
\centering
\setlength{\tabcolsep}{3.0pt}
\renewcommand{\arraystretch}{1.15}
\resizebox{\textwidth}{!}{
\begin{tabular}{lcccccccccccc|cc}
\toprule
\multirow{2}{*}{\textbf{Methods}}
& \multicolumn{4}{c}{\textbf{Arithmetic}}
& \multicolumn{4}{c}{\textbf{OpenBookQA}}
& \multicolumn{4}{c|}{\textbf{BoolQ}}
& \multirow{2}{*}{\shortstack{\textbf{Mean}\\\textbf{Trade.} $\uparrow$}}
& \multirow{2}{*}{\shortstack{\textbf{Gain}\\\textbf{vs. Best} $\uparrow$}} \\
\cmidrule(lr){2-5}
\cmidrule(lr){6-9}
\cmidrule(lr){10-13}
& \textbf{F Avg.} $\downarrow$
& \textbf{F@20} $\downarrow$
& \textbf{R Avg.} $\uparrow$
& \textbf{Trade.} $\uparrow$
& \textbf{F Avg.} $\downarrow$
& \textbf{F@20} $\downarrow$
& \textbf{R Avg.} $\uparrow$
& \textbf{Trade.} $\uparrow$
& \textbf{F Avg.} $\downarrow$
& \textbf{F@20} $\downarrow$
& \textbf{R Avg.} $\uparrow$
& \textbf{Trade.} $\uparrow$
& & \\
\midrule

GA (ACL'23)
& 63.25 & 08.00 & 58.92 & 45.27
& 33.25 & 34.00 & 71.17 & 68.89
& 80.00 & 80.00 & 65.92 & 30.69
& \cellcolor{LightBlue}48.28
& \cellcolor{LightBlue}-2.76 \\

RMU (ICML'24)
& 89.50 & 90.00 & \textbf{66.42} & 18.13
& 35.75 & 35.00 & 71.50 & 67.68
& 80.00 & 80.00 & \textbf{66.17} & 30.72
& \cellcolor{LightBlue}38.84
& \cellcolor{LightBlue}-12.19 \\

SimNPO (NeurIPS'25)
& 81.25 & 80.00 & 65.67 & 29.17
& 35.50 & 36.00 & 70.25 & 67.25
& 81.25 & 80.00 & 65.50 & 29.15
& \cellcolor{LightBlue}41.86
& \cellcolor{LightBlue}-9.19 \\

LLM-Eraser (KDD'25)
& \textbf{13.25} & 00.00 & 34.25 & 49.11
& 35.75 & 38.00 & 53.42 & 58.33
& 64.75 & 65.00 & 54.17 & 42.71
& \cellcolor{LightBlue}50.05
& \cellcolor{LightBlue}-0.99 \\

O3 (ICLR'25)
& 76.75 & 60.00 & 62.67 & 33.92
& 36.75 & 38.00 & 70.00 & 66.45
& 78.75 & 69.00 & 65.75 & 32.12
& \cellcolor{LightBlue}44.16
& \cellcolor{LightBlue}-6.88 \\

ASU (ICLR'26)
& 58.00 & 36.00 & 48.25 & 44.91
& 36.00 & 35.00 & \textbf{72.17} & 67.84
& 70.25 & 65.00 & 62.75 & 40.36
& \cellcolor{LightBlue}51.04
& \cellcolor{LightBlue}0.00 \\

\hline

\textbf{TFR-Net (Ours)}
& 23.75
& \textbf{00.00}
& 51.92
& \textbf{61.77}
& \textbf{30.00}
& \textbf{27.00}
& 70.25
& \textbf{70.12}
& \textbf{63.25}
& \textbf{55.00}
& 62.42
& \textbf{46.26}
& \cellcolor{LightBlue}\textbf{59.39}
& \cellcolor{LightBlue}\textbf{+8.35} \\

\bottomrule
\end{tabular}
}
\caption{
Main results (\%).
F Avg.\ and F@20 denote average and final forget-set accuracy, where lower is better.
R Avg.\ is retained utility averaged over three evaluation sets and four checkpoints.
Mean Trade.\ averages Trade-off across the three forget sets.
Gain vs.\ Best is the signed percentage-point difference from the best baseline in Mean Trade.
}
\label{tab:compact_main_results}
\end{table*}

\subsection{Retain-Guarded Capacity Recovery}
This subsection recovers usable model capacity from dormant channels after trajectory-aware suppression. TFR-Net reactivates dormant channels with strong contributions to retained utility under constraints on current and historical forget risk.

After trajectory-aware channel suppression, TFR-Net obtains a pre-recovery mask as follows:
\begin{equation}
\widehat{m}_t(c)
=
m_{t-1}(c)\alpha_t(c),
\label{eq:pre_recovery_mask}
\end{equation}
where $\alpha_t(c)$ is the suppression factor defined in the previous subsection. The value $\widehat{m}_t(c)$ represents the channel state after trajectory-aware suppression but before capacity recovery.

Because pre-mask activations keep channel scores informative under dormancy, TFR-Net admits only those with low current and historical forget risk into the recovery candidate pool:
\begin{equation}
\mathcal{Q}_t
=
\{c\mid \widehat{m}_t(c)\leq\tau_d,\ \chi_t(c)\leq\tau_f,\ \Gamma_{t-1}(c)\leq\tau_h\},
\label{eq:recovery_candidate_pool}
\end{equation}
where $\mathcal{Q}_t$ is the recovery candidate pool, $\tau_d$ is the dormant threshold, $\tau_f$ controls current forget-risk filtering, and $\tau_h$ thresholds the previous-request trajectory-aware channel priority. This constraint prevents channels with high current or historical risk from immediate recovery.

For each candidate channel, TFR-Net evaluates its recovery value by combining retained utility and forget-risk penalties. The recovery score is defined as follows:
\begin{equation}
\upsilon_t(c)
=
\varpi_u\bar{\psi}_t^{r}(c)
-
\varpi_c\chi_t(c)
-
\varpi_h\Gamma_{t-1}(c)
-
\varpi_\omega\omega_{t-1}(c),
\label{eq:recovery_score}
\end{equation}
where $\bar{\psi}_t^{r}(c)$ is the standardized retain score, $\chi_t(c)$ is the current-request channel risk, $\Gamma_{t-1}(c)$ is the previous-request trajectory-aware channel priority, and $\omega_{t-1}(c)$ records repeated high-risk occurrences before request $t$. The coefficients $\varpi_u$, $\varpi_c$, $\varpi_h$, and $\varpi_\omega$ weight the retain contribution and the three risk penalties.

TFR-Net selects channels whose recovery scores exceed $\tau_g$. The selected recovery set is written as follows:
\begin{equation}
\mathcal{U}_t
=
\{c\mid c\in\mathcal{Q}_t,\ \upsilon_t(c)>\tau_g,\ c\notin\mathcal{P}_t\},
\label{eq:recovery_selection}
\end{equation}
where $\mathcal{U}_t$ denotes the selected recovery channels, $\tau_g$ is the recovery threshold, and $\mathcal{P}_t$ is the high-priority suppression set selected in the previous subsection. The condition $c\notin\mathcal{P}_t$ avoids suppressing and restoring the same channel in one request.

The recovery increment is assigned according to the selected recovery set as follows:
\begin{equation}
\eta_t(c)
=
\delta_{\mathrm{rec}}\big(1-\widehat{m}_t(c)\big)\mathbb{I}\big[c\in\mathcal{U}_t\big],
\label{eq:recovery_increment}
\end{equation}
where $\delta_{\mathrm{rec}}$ controls the recovery strength. This design restores only a fraction of the missing mask value, making capacity recovery gradual rather than fully reactivating dormant channels in one step. When $c=(l,g,j)$, $\eta_t(c)$ corresponds to $\eta_t^{l,g,j}$ in the forget-recover transition.

After suppression and recovery, TFR-Net obtains a tentative controller
state $\widetilde{\mathcal{X}}_t$, whose mask component is
$\widetilde{\mathbf{m}}_t$. The state also contains the updated trajectory
memory, suppression history, and controller variables produced during the
current request. Let $G_{t-1}^{r}$ and $\widetilde{G}_t^{r}$ denote the retain losses before and after the tentative update, respectively. The acceptance criterion is:
\begin{equation}
\widetilde{G}_t^{r}
\leq
\bigl(1+\epsilon_r^{\mathrm{rel}}\bigr)G_{t-1}^{r}
+
\epsilon_r^{\mathrm{abs}},
\label{eq:retain_guard}
\end{equation}
where $\epsilon_r^{\mathrm{rel}}$ and $\epsilon_r^{\mathrm{abs}}$ are the
relative and absolute retain-loss tolerances.

The complete controller state is then accepted or restored jointly:
\begin{equation}
\mathcal{X}_t
=
\begin{cases}
\widetilde{\mathcal{X}}_t,
& \text{if Eq.~\eqref{eq:retain_guard} holds},\\
\mathcal{X}_{t-1},
& \text{otherwise}.
\end{cases}
\label{eq:atomic_state_commit}
\end{equation}
Therefore, a rejected proposal restores the full pre-request state rather than reverting recovery alone. All tentative updates to the mask, trajectory memory, and suppression history are discarded together. This atomic rollback prevents partial structural updates and keeps the accepted mask consistent with its associated trajectory memory.

\section{Experiments}
\subsection{Experimental Settings}
\textbf{Baselines.}
We compare TFR-Net with six representative LLM unlearning baselines. GA~\cite{JangYY2023} is a direct optimization baseline that maximizes the forget loss. RMU~\cite{PaGY2024} suppresses target knowledge by steering forget representations away from their original activations. SimNPO~\cite{FaLL2026} performs reference-free negative preference optimization for stable unlearning. LLM-Eraser~\cite{ZhaZZ2025} removes unlearning-relevant neurons through selective pruning. O3~\cite{GaoWD2025} addresses sequential forget requests with continual unlearning mechanisms. ASU~\cite{ZadZL2026} weakens memorized associations via attention-smoothing self-distillation. 

\noindent \textbf{Datasets.}
We evaluate TFR-Net on BoolQ~\cite{ClLC2019}, OpenBookQA~\cite{MiCK2018}, Arithmetic~\cite{BrMR2020}, and TOFU~\cite{MaiFSi2024}. For the first three datasets, each experiment forgets one task while evaluating retention on the others and a held-out same-task split. On TOFU, we use the official \texttt{forget10}/\texttt{retain90} protocol with a retain-only oracle to assess forgetting, utility, and privacy. We use candidate ranking for BoolQ and OpenBookQA, direct generation for Arithmetic, and the official generation- and likelihood-based evaluation for TOFU.

\noindent \paragraph{\textbf{Evaluation Metrics}}
Forget-set accuracy is averaged over Req.~5, 10, 15, and 20, with the final accuracy reported at Req.~20. Retain-set accuracy is averaged over the same-task split and the other two datasets at these four checkpoints. Trade-off is the harmonic mean of forgetting effectiveness and retained utility, and the Mean Trade-off averages the results across the three forget streams. For TOFU, we follow the official evaluation protocol~\cite{MaiFSi2024} and report membership inference attack gap, privacy leakage, and model utility.

\noindent \paragraph{\textbf{Implementation Details.}}
All experiments are conducted on NVIDIA RTX PRO 4500 GPUs. We use LLaMA-7B as the backbone and perform continual unlearning over 20 forget requests. Each request contains 8 forget examples, with sequences truncated to 256 tokens. The maximum global pruning budget is set to 128 channels per request according to the dataset, the recovery ratio is 0.5, the maximum mask sparsity is 0.03--0.04, and the retain-safe tolerance is $\epsilon_r=0.05$. We evaluate the model before unlearning and every 5 requests thereafter using 100 examples per task.

\subsection{Main Results}

\begin{table}[t]
\centering
\tiny
\setlength{\tabcolsep}{0.8pt}
\renewcommand{\arraystretch}{0.72}
\resizebox{\columnwidth}{!}{%
\begin{tabular}{llcc cc cc cc}
\toprule
\multirow{2}{*}{Dataset}
& \multirow{2}{*}{Variant}
& \multicolumn{2}{c}{Forget $\downarrow$}
& \multicolumn{6}{c}{Retain $\uparrow$} \\
\cmidrule(lr){3-4}
\cmidrule(lr){5-10}
&
&
\multicolumn{2}{c}{--}
& \multicolumn{2}{c}{BoolQ}
& \multicolumn{2}{c}{OBQA}
& \multicolumn{2}{c}{Arith.} \\
\cmidrule(lr){3-4}
\cmidrule(lr){5-6}
\cmidrule(lr){7-8}
\cmidrule(lr){9-10}
&
& R5 & R20
& R5 & R20
& R5 & R20
& R5 & R20 \\
\midrule

\multirow{4}{*}{Arith.}
& w/o recover
& 66.0 & 00.0
& 78.0 & 75.0
& 38.0 & 38.0
& 61.0 & 00.0 \\
& w/o Retain
& 66.0 & 00.0
& 78.0 & 78.0
& 38.0 & 35.0
& 61.0 & 00.0 \\
& w/o Hist.-risk
& 66.0 & 04.0
& 78.0 & 80.0
& 38.0 & 38.0
& 61.0 & 11.0 \\

\rowcolor{LightBlue}
& \textbf{TFR-Net}
& 80.0 & \textbf{00.0}
& \textbf{82.0} & \textbf{83.0}
& 37.0 & \textbf{36.0}
& \textbf{76.0} & \textbf{25.0} \\

\midrule

\multirow{4}{*}{OBQA}
& w/o recover
& 32.0 & 28.0
& 82.0 & 79.0
& 45.0 & 36.0
& 89.0 & 89.0 \\
& w/o Retain
& 32.0 & 28.0
& 82.0 & 80.0
& 45.0 & 38.0
& 89.0 & 90.0 \\
& w/o Hist.-risk
& 32.0 & 29.0
& 82.0 & 78.0
& 45.0 & \textbf{40.0}
& 89.0 & \textbf{91.0} \\

\rowcolor{LightBlue}
& \textbf{TFR-Net}
& \textbf{32.0} & \textbf{27.0}
& 81.0 & \textbf{78.0}
& \textbf{45.0} & 38.0
& \textbf{89.0} & 90.0 \\

\midrule

\multirow{4}{*}{BoolQ}
& w/o recover
& 80.0 & 64.0
& 72.0 & 61.0
& 36.0 & 32.0
& 90.0 & 67.0 \\
& w/o Retain
& 80.0 & 70.0
& 72.0 & \textbf{63.0}
& 36.0 & 36.0
& 90.0 & 84.0 \\
& w/o Hist.-risk
& 80.0 & 59.0
& 72.0 & 53.0
& 36.0 & 36.0
& 90.0 & 78.0 \\

\rowcolor{LightBlue}
& \textbf{TFR-Net}
& \textbf{74.0} & \textbf{55.0}
& \textbf{72.0} & 53.0
& \textbf{36.0} & \textbf{37.0}
& \textbf{91.0} & \textbf{89.0} \\

\bottomrule
\end{tabular}%
}
\caption{
TFR-Net ablation across three datasets.
R5 and R20 denote Req.~5 and Req.~20.
w/o Recovery, w/o Retain, and w/o Hist.-risk
denote variants without capacity recovery, retain guidance, and
historical-risk screening, respectively.
}
\label{tab:ablation_study}
\end{table}

Table~\ref{tab:compact_main_results} compares average and final forget-set accuracy, retained utility, and Trade-off. TFR-Net achieves the highest Trade-off on all three forget sets and improves Mean Trade.\ over the best baseline by 8.34 percentage points. Its consistent advantage across datasets indicates a stable balance between unlearning effectiveness and retained utility.

\noindent\paragraph{\textbf{General unlearning baselines.}} GA forgets effectively only on Arithmetic, whereas RMU and SimNPO preserve retained utility but retain substantial target knowledge. LLM-Eraser achieves aggressive forgetting on Arithmetic, yet its utility degradation and weaker results on OpenBookQA and BoolQ expose the limitations of localization in a single model state. TFR-Net instead uses channel-level risk trajectories to suppress persistent target-related channels while avoiding transient hotspots, which is consistent with its stronger forgetting and competitive retained utility across all three streams.

\noindent\paragraph{\textbf{Continual unlearning baselines.}} O3 generally preserves retained utility but leaves considerable residual target knowledge, while ASU improves forgetting at a greater utility cost, particularly on Arithmetic. Their dataset-dependent behavior suggests that stabilizing successive updates alone does not address changing target relevance or cumulative capacity loss. TFR-Net tracks pathway evolution across requests and reactivates low-risk dormant channels under a retained-utility guard, consistent with its more uniform forgetting--utility balance across all three streams.

\begin{figure}[t]
\centering
\includegraphics[width=\linewidth]{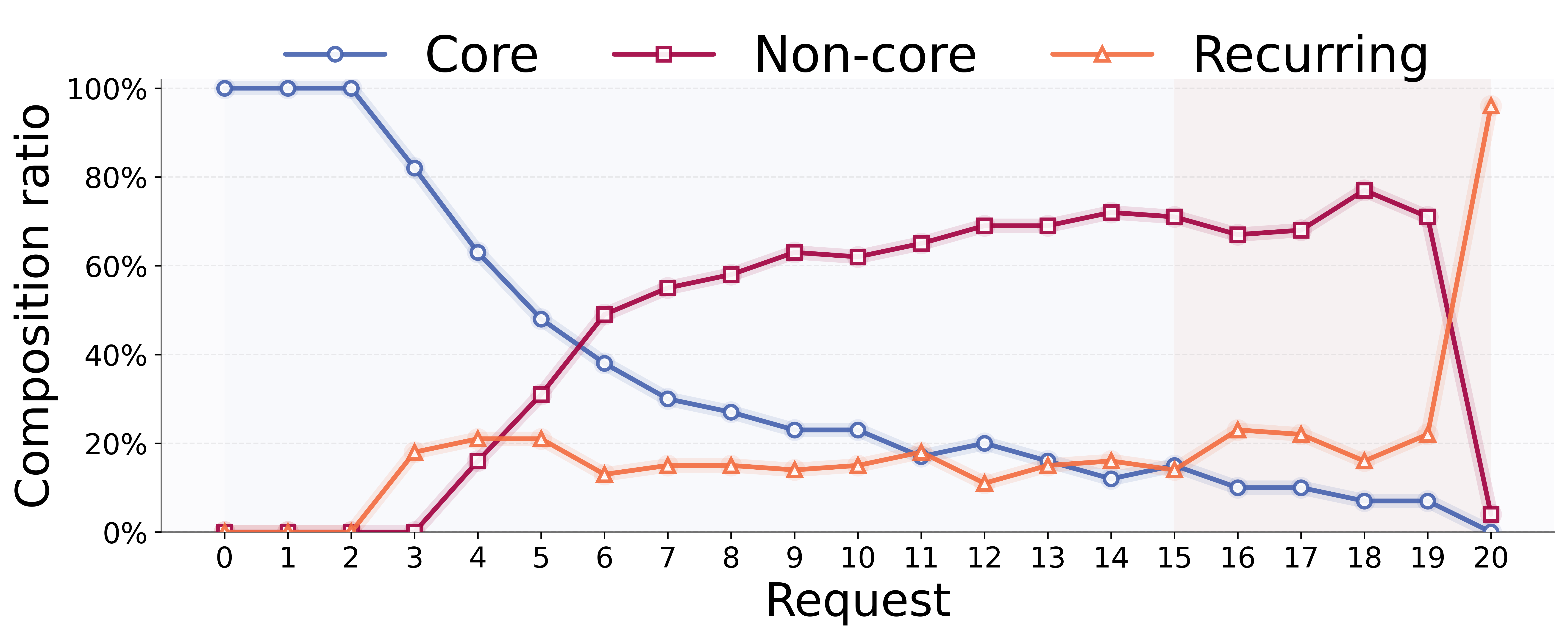}
\caption{
Cross-request composition of the Top-$K$ channels ranked by target relevance
on BoolQ.
Core denotes channels in the Req.~1 Top-$K$ set;
Non-core denotes channels first observed in the current request;
Recurring denotes non-core channels observed in earlier requests.
}
\label{fig:relevance_boolq}
\end{figure}
\begin{figure}[t]
\centering
\includegraphics[width=\linewidth]{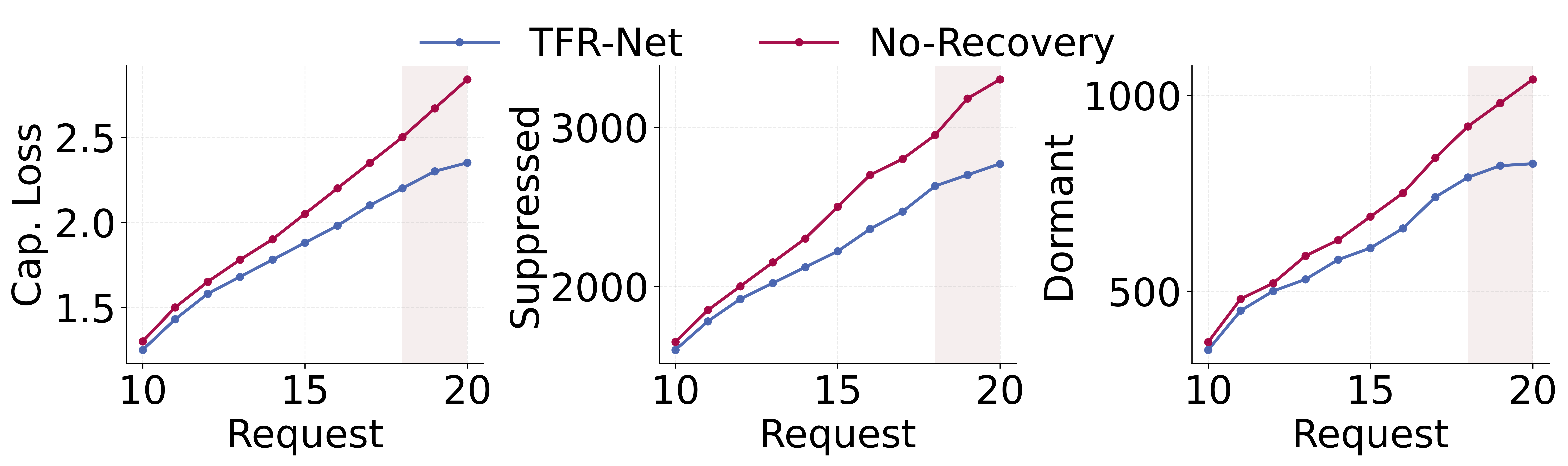}
\caption{Capacity dynamics on BoolQ.
Capacity Loss is
$100(1-\mathrm{Cap}(\mathbf m_t))$ in percentage points;
Suppressed and Dormant are channel counts.}
\label{fig:capacity_boolq_f}
\end{figure}

\subsection{Ablation Study}
Table~\ref{tab:ablation_study} shows that some ablations match or exceed TFR-Net on individual metrics at R5, whereas their limitations become more evident at R20, indicating that the three components primarily improve stability across successive forget requests. Without Recovery, final forgetting is preserved on Arithmetic, but retained utility declines substantially in several same-task and cross-task evaluations, particularly when forgetting Arithmetic and BoolQ. This pattern is consistent with capacity recovery mitigating the cumulative loss of usable channels rather than optimizing every isolated metric. Without Retain Guidance, same-task retained utility collapses on Arithmetic, while substantially more target knowledge remains on BoolQ, showing that channel updates without retain guidance can favor either forgetting or utility at the expense of the other. Without Historical Risk, final forget-set accuracy is consistently higher on all three forget sets, with additional utility degradation in several settings. This result indicates that current-request risk alone is insufficient to constrain channels associated with earlier forget requests. Although no configuration dominates every entry, TFR-Net avoids these recurring endpoint failures and provides the most consistent trade-off between unlearning effectiveness and retained utility.

\subsection{Privacy--Utility Evaluation on TOFU}
\begin{table}[t]
\centering
\tiny
\setlength{\tabcolsep}{3pt}
\renewcommand{\arraystretch}{1.15}
\resizebox{\columnwidth}{!}{%
\begin{tabular}{lccc}
\toprule
\textbf{Methods}
&
\textbf{MIA Gap $\downarrow$}
&
\textbf{PrivLeak $\downarrow$}
&
\textbf{Model Utility $\uparrow$}
\\
\midrule

Full SFT
&0.0620
&28.13
&0.3414
\\

GA
&0.0622
&28.10
&\textbf{0.3415}
\\

SimNPO
&0.0619
&28.05
&0.3412
\\

RMU
&0.0630
&28.23
&0.3405
\\

LLM-Eraser
&0.1108
&43.37
&0.0000
\\

O3
&0.0622
&28.12
&0.3411
\\

ASU
&0.0623
&28.02
&0.3412
\\
\hline
\rowcolor{LightBlue}
\textbf{TFR-Net (Ours)}
&\textbf{0.0587}
&\textbf{27.29}
&0.3409
\\

\bottomrule
\end{tabular}
}
\caption{
Privacy--utility comparison on TOFU.
MIA Gap is the mean of $|\mathrm{AUC}-0.5|$ across four black-box
membership inference attacks, and PrivLeak is the absolute deviation from the retain-only oracle. Full SFT is the original model before unlearning.
}
\label{tab:privacy_main}
\end{table}
As shown in Table~\ref{tab:privacy_main}, TFR-Net achieves the lowest MIA Gap and PrivLeak while retaining 99.85\% of the Model Utility of Full SFT. Relative to Full SFT, it reduces the two privacy risks by 5.3\% and 3.0\%, respectively, and also outperforms SimNPO and ASU, the strongest competing baselines on the corresponding metrics. Its consistent advantage on both privacy indicators demonstrates more effective removal of membership signals. Most utility-preserving methods remain close to Full SFT on both metrics, indicating limited privacy improvement despite preserving utility. LLM-Eraser instead increases both privacy risks and reduces Model Utility to zero, showing that broad capability degradation does not ensure effective privacy unlearning. TFR-Net is therefore the only method that achieves the best result on both privacy metrics while keeping Model Utility essentially unchanged. This behavior is consistent with trajectory-aware suppression distinguishing persistent target-related channels from transient hotspots, while retain-guarded capacity recovery reactivates dormant channels with strong contributions to retained utility and low current and historical forget risk.

\subsection{Cross-Request Structural Analysis}
Figures~\ref{fig:relevance_boolq} and~\ref{fig:capacity_boolq_f} examine the two structural phenomena underlying TFR-Net on BoolQ: the redistribution of target relevance and the cumulative loss of usable model capacity.

\noindent\paragraph{\textbf{Target relevance redistribution.}} Figure~\ref{fig:relevance_boolq} shows that the Initial Core decreases from the entire Top-$K$ set to only a small fraction, while New channels dominate most later requests and Recurring channels repeatedly reappear. Target relevance is therefore associated with a changing set of channels rather than a fixed initial core. Moreover, a score observed in one request does not reveal whether a channel will remain target-related or appear only transiently. This observation motivates the channel-level risk trajectories in TFR-Net, which prioritize persistent target-related channels for suppression while avoiding transient hotspots.

\noindent\paragraph{\textbf{Cumulative capacity loss.}} Figure~\ref{fig:capacity_boolq_f} shows that capacity loss, channel suppression, and dormancy increase under both configurations, confirming that repeated suppression progressively reduces usable model capacity. The variant without recovery consistently accumulates larger values, with the differences becoming more pronounced in later requests. This result supports retain-guarded capacity recovery as a complement to trajectory-aware suppression: TFR-Net reactivates dormant channels with strong contributions to retained utility and low current and historical forget risk, thereby mitigating cumulative capacity loss without indiscriminate channel restoration.

\section{Conclusion}
Continual unlearning is fundamentally a long-horizon problem of preserving prior forgetting without exhausting the model capacity needed for retained utility. Our analysis reveals two coupled structural effects: target-related computation shifts across channels, while repeated suppression progressively reduces usable model capacity. TFR-Net addresses pathway evolution through channel-level risk trajectories that distinguish persistent target-related channels from transient hotspots. It addresses capacity loss through retain-guarded recovery, which selectively reactivates dormant channels with strong contributions to retained utility and low current and historical forget risk. A recovered state is accepted only when retained-utility degradation remains within a predefined tolerance. Across four datasets, TFR-Net consistently improves the forgetting--utility balance and achieves the highest Trade-off on all three continual unlearning streams. Ablation and structural analyses further demonstrate that tracking pathway evolution and preserving usable model capacity are both essential for reliable continual unlearning.



\bibliography{aaai2027}


\end{document}